\documentclass[letterpaper, 10 pt, conference]{ieeeconf}  % Comment this line out if you need a4paper

\usepackage{amsmath}
\usepackage{amssymb}   %（推荐，一起加）
\usepackage{graphicx}
\usepackage{cite}
\usepackage{algorithm}
\usepackage{algpseudocode}
\usepackage{booktabs}
\usepackage[table]{xcolor}
\usepackage{graphicx}
\usepackage{booktabs}   % 更好看的横线
\usepackage{multirow}   % 跨行单元格
\usepackage{siunitx}    % 数字对齐（可选但强烈推荐）
\IEEEoverridecommandlockouts                              % This command is only needed if 
\title{\LARGE \bf
Anticipatory Risk-Guided Reinforcement Learning for Safe Flight Through Dynamic Clutter
}

\author{
Yuchao Mei$^{*}$, Guohao Zhang$^{*}$, Luxia Ai, Haopeng Chen, and Wenbing Tao$^{\dagger}$%
\thanks{%
$^{*}$Yuchao Mei and Guohao Zhang contributed equally to this work.
$^{\dagger}$Wenbing Tao is the corresponding author.%
}
\thanks{%
This work was supported by the National Natural Science Foundation of China under Grant No.~6257074579 and the Hubei Provincial Science and Technology Plan Project under Grant No.~2025BEB006.%
}%
\thanks{%
Yuchao Mei, Guohao Zhang, Luxia Ai, Haopeng Chen, and Wenbing Tao are with the National Key Laboratory of Science and Technology on Multi-spectral Information Processing, School of Artificial Intelligence and Automation, Huazhong University of Science and Technology, Wuhan 430074, China.
E-mail: \{yuchaomei, wenbingtao\}@hust.edu.cn.
}%
}

\begin{document}

\maketitle
\thispagestyle{empty}
\pagestyle{empty}

%%%%%%%%%%%%%%%%%%%%%%%%%%%%%%%%%%%%%%%%%%%%%%%%%%%%%%%%%%%%%%%%%%%%%%%%%%%%%%%%
%%%%%%%%%%%%%%%%%%%%%%%%%%%%%%%%%%%%%%%%%%%%%%%%%%%%%%%%%%%%%%%%%%%%%%%%%%%%%%%%
\begin{abstract}
Safe quadrotor navigation in cluttered and dynamic environments depends not only on instantaneous geometric perception, but more critically on anticipating collision risks induced by relative motion. Conventional modular pipelines frequently suffer from perception latency, while end-to-end learning methods relying on implicit scalar rewards often struggle to extract reliable spatio-temporal features without physics-grounded supervision. To address this, we propose an anticipatory risk-guided reinforcement learning framework. Leveraging privileged simulator states, we construct a directionally aligned future collision risk map based on the Closest Point of Approach (CPA). Through an asymmetric actor-critic architecture, the network is trained to self-predict this structured risk, which explicitly guides the visual policy during deployment. A lightweight spatio-temporal encoder extracts motion cues directly from onboard depth sequences, bypassing explicit object tracking or optical flow estimation. Extensive simulated and real-world experiments demonstrate that our method effectively improves safety margins and flight efficiency in dense dynamic clutters compared to existing baselines. Furthermore, the learned policy achieves robust zero-shot Sim-to-Real transfer on a physical quadrotor, relying purely on abstracted spatio-temporal depth sequences and its self-predicted risk priors, validating the effectiveness of our approach and its robust generalization from simulation to reality.
\end{abstract}
%%%%%%%%%%%%%%%%%%%%%%%%%%%%%%%%%%%%%%%%%%%%%%%%%%%%%%%%%%%%%%%%%%%%%%%%%%%%%%%%
%%%%%%%%%%%%%%%%%%%%%%%%%%%%%%%%%%%%%%%%%%%%%%%%%%%%%%%%%%%%%%%%%%%%%%%%%%%%%%%%
\section{Introduction}
Autonomous navigation of micro aerial vehicles (MAVs) in cluttered, dynamic environments remains a grand challenge \cite{wang2021autonomous, lin2020robust}. Unlike static scenarios, safe dynamic navigation requires not only instantaneous geometric clearance but also the explicit anticipation of relative motion. Because dynamic obstacles introduce critical temporal dependencies, their latent collision risks cannot be reliably captured by a single spatial snapshot. Therefore, enabling MAVs to extract these motion cues and execute proactive avoidance using only continuous onboard sensing is essential for robust real-world deployment.

\begin{figure}[htbp]
    \centering
    \includegraphics[width=1.0\linewidth]{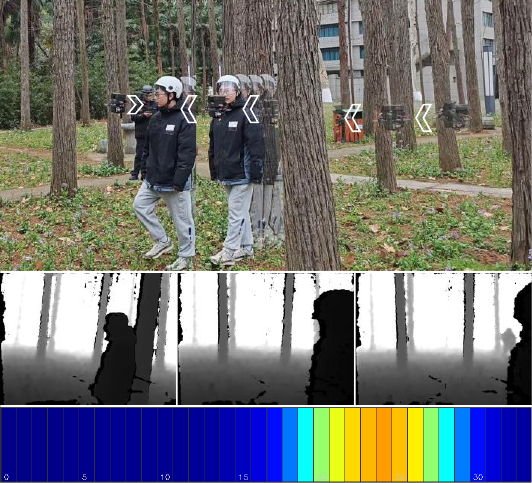}
    \caption{Real-world demonstration of the proposed anticipatory navigation framework in a static-dynamic environment. \textbf{Top:} Third-person view of the quadrotor avoiding pedestrians and static obstacles. \textbf{Middle:} Sequence of raw ego-centric depth images captured onboard. \textbf{Bottom:} The predicted 1D directional future collision risk map.}
    \label{fig:teaser}
\end{figure}

Traditional modular pipelines typically rely on explicit object tracking to anticipate dynamic threats. However, these methods frequently suffer from computational latency and error accumulation under severe occlusion \cite{lu2022perception, xu2023vision, lu2025fapp}. As a promising alternative, end-to-end reinforcement learning (RL) circumvents these explicit state estimation bottlenecks by extracting motion cues directly from sequential visual data \cite{loquercio2021learning, hoeller2021learning, fan2025flying}. However, a critical limitation of existing end-to-end RL methods is their reliance on implicit guidance via aggregated, task-level scalar rewards (e.g., distance progress and collision penalties) \cite{xu2025flow, zhao2024learning}. Since such scalar feedback lacks explicit spatial and temporal structure, it is fundamentally insufficient for capturing complex obstacle dynamics. Maximizing task-level rewards often incentivizes aggressive, last-second reactive behaviors that trade safety margins for speed. Without explicit supervision on ``where'' and ``when'' a collision might occur, feature extractors struggle to prioritize safety-critical dynamic cues, resulting in brittle navigation vulnerable to sudden dynamic threats.

This work directly addresses these limitations by introducing an end-to-end risk-guided RL framework that embeds explicit geometric motion priors into visual learning, prioritizing absolute safety over reckless agility. Our core insight is to utilize privileged simulator information---the ground-truth motion states of obstacles---to construct a directionally aligned Future Collision Risk Map based on the Closest Point of Approach (CPA). This risk map explicitly encodes the temporal urgency and spatial proximity of potential threats. Leveraging an asymmetric actor-critic architecture, we apply this privileged representation as a dense supervisory signal to explicitly guide the spatio-temporal feature encoder during training. This risk distillation forces the visual backbone to extract reliable motion cues for precise collision anticipation, yielding a deployable policy that operates purely on onboard depth sequences. Consequently, the proposed approach surpasses implicit learning baselines by executing proactive spatial detours in advance, which maintains safer clearance distances rather than relying on aggressive, late-reactive maneuvers. Furthermore, the learned policy achieves robust zero-shot Sim-to-Real transfer without fine-tuning, validating its strong generalization capability.

To summarize, this work makes the following contributions:
\begin{itemize}
    \item \textbf{Anticipatory Risk Representation:} We formulate a directionally-aligned future collision risk map based on CPA analysis. This structured representation explicitly encodes temporal urgency and spatial proximity, providing physics-grounded supervision for dynamic avoidance.
    
    \item \textbf{Asymmetric Risk-Guided RL Framework:} We develop an actor-critic architecture integrating a lightweight, tracking-free spatio-temporal depth encoder. By extracting dynamic cues directly from geometry-consistent observation sequences, the policy explicitly self-predicts collision risks to guide safe navigation, bypassing the latency of explicit object tracking.
    
    \item \textbf{Robust Zero-Shot Sim-to-Real Transfer:} We validate the proposed approach through comprehensive simulation and physical quadrotor experiments. Our method demonstrates superior safety margins, reliable anticipatory avoidance, and robust zero-shot deployment in highly dynamic real-world clutters.
\end{itemize}
%%%%%%%%%%%%%%%%%%%%%%%%%%%%%%%%%%%%%%%%%%%%%%%%%%%%%%%%%%%%%%%%%%%%%%%%%%%%%%%%
\section{Related Work}

\subsection{Model-based Navigation in Dynamic Environments}
Traditional model-based approaches typically follow a ``perception-modeling-planning'' pipeline, explicitly estimating the geometry and motion states of obstacles. 
For perception, Wang \textit{et al.} \cite{wang2021autonomous} modeled dynamic obstacles as ellipsoids, while Lu \textit{et al.} \cite{lu2022perception} combined lightweight detection with 3D-SORT tracking. 
More recently, Lu \textit{et al.} \cite{lu2025fapp} introduced the FAPP framework, employing covariance adaptation to accurately predict the motion of multiple dynamic targets. 
To handle arbitrary obstacle shapes, Chen \textit{et al.} \cite{chen2023risk} designed a Dual-Structure Particle (DSP) dynamic occupancy map. Regarding planning and control, handling uncertainty is paramount. 
Lin \textit{et al.} \cite{lin2020robust} and Xu \textit{et al.} \cite{xu2022dpmpc} utilized Chance-Constrained Model Predictive Control (CC-MPC) to explicitly account for sensing noise and prediction errors. 
To improve real-time performance, Xu \textit{et al.} \cite{xu2023vision} integrated gradient-based B-spline optimization with a Receding Horizon Distance Field. 
Furthermore, to mitigate overly conservative behaviors, Xu \textit{et al.} \cite{xu2025intent} integrated intent prediction into MPC to forecast obstacle trajectories, while Liu \textit{et al.} \cite{liu2025dynamic} combined Time-Adaptive MPC with Dynamic Control Barrier Functions (D-CBF) to address the ``freezing robot problem''. 
Despite their capabilities, these methods rely on hand-crafted representations and simplified motion models. In highly stochastic environments, prediction errors often lead to overly conservative behaviors or planning failures.

\subsection{Learning-based Navigation in Dynamic Environments}
Learning-based approaches map sensor observations directly to control commands, significantly reducing decision latency. Loquercio \textit{et al.} \cite{loquercio2021learning} and Tordesillas \textit{et al.} \cite{tordesillas2023deep} employed imitation learning (IL) to train visual policies for high-speed flight and high-frequency replanning. However, IL's generalization remains limited by expert data coverage. Within reinforcement learning (RL), Yang \textit{et al.} \cite{yang2020autonomous} applied Double Deep Q-Networks for dynamic avoidance, while Wang \textit{et al.} \cite{wang2023curriculum} employed curriculum learning to handle movable obstacles. To incorporate explicit avoidance priors, Xie \textit{et al.} \cite{xie2023drl} introduced Velocity Obstacles, and Xu \textit{et al.} \cite{xu2025navrl} integrated explicit dynamic object detection. Yet, these methods typically rely on independent detection modules, increasing susceptibility to sensing errors and system latency.

Recent studies have increasingly explored implicit dynamic representations to bypass explicit object tracking. For instance, recent works \cite{fan2025flying, xu2025flying} mapped compressed multi-frame point clouds to control commands, whereas Hoeller \textit{et al.} \cite{hoeller2021learning} and de Heuvel \textit{et al.} \cite{de2024spatiotemporal} extracted spatiotemporal representations from sequential sensor data. Furthermore, Xu \textit{et al.} \cite{xu2025flow} introduced point flow as a dynamic scene representation. While effective, these approaches guide representation learning indirectly through scalar, task-level rewards. Since scalar feedback lacks explicit spatial and temporal supervision, these networks often struggle to encode the complex kinematics critical for anticipatory avoidance. Although learning speed adaptation \cite{zhao2024learning} can mitigate overly aggressive maneuvers, policies lacking geometric priors frequently exhibit purely reactive behaviors.

Motivated by these limitations, we propose an end-to-end risk-guided RL framework that explicitly models short-horizon collision risk. By supervising a spatiotemporal depth encoder with a privileged ground-truth risk map during training, our method extracts motion-aligned features. Consequently, the deployed network can explicitly self-predict these risks, enabling the policy to make proactive, safety-first dynamic navigation decisions directly from visual observations.
%%%%%%%%%%%%%%%%%%%%%%%%%%%%%%%%%%%%%%%%%%%%%%%%%%%%%%%%%%%%%%%%%%%%%%%%%%%%%%%%
%%%%%%%%%%%%%%%%%%%%%%%%%%%%%%%%%%%%%%%%%%%%%%%%%%%%%%%%%%%%%%%%%%%%%%%%%%%%%%%%
\begin{figure*}[htbp]
    \centering
    % \vspace*{4mm}
    \includegraphics[width=\textwidth, trim=1 1 1 1, clip]{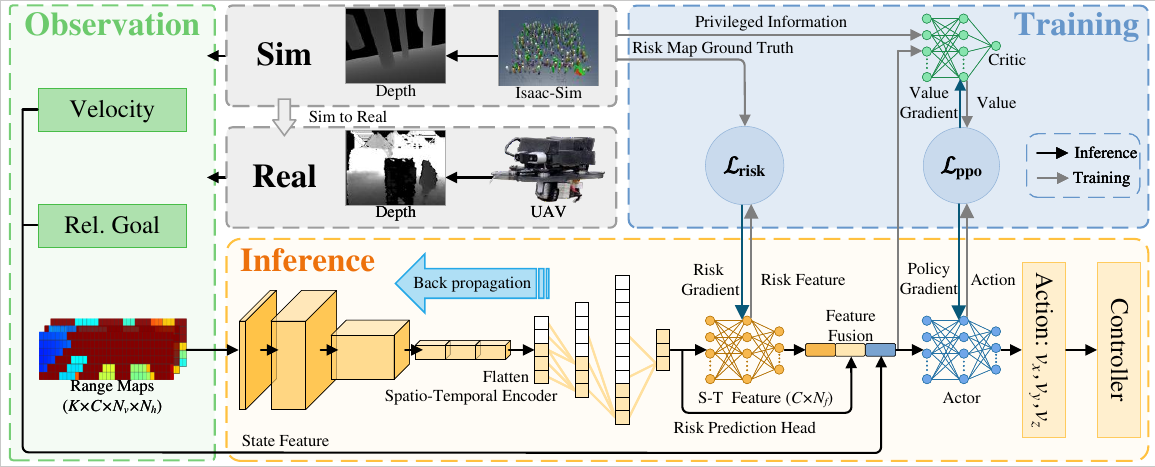}
    \caption{Overview of the proposed anticipatory risk-guided navigation framework. \textbf{Observation:} The policy takes proprioceptive states (velocity and relative goal position) and a temporal sequence of range maps as inputs. \textbf{Training:} An asymmetric actor-critic architecture is employed. Privileged ground-truth obstacle states from the simulator are utilized to construct a CPA-based future collision risk map. This map serves as a dense supervisory signal ($\mathcal{L}_{\text{risk}}$) for the auxiliary prediction head, explicitly guiding the Spatio-Temporal Encoder to learn motion-aware representations via backpropagation. Meanwhile, the critic leverages full privileged information to stabilize policy updates. \textbf{Inference:} Privileged modules are seamlessly detached. The vision-based Actor outputs continuous velocity commands relying solely on the fused representation of the encoded spatio-temporal features, its self-predicted risk, and the proprioceptive state features, enabling robust, tracking-free dynamic avoidance in the real world.}
    \label{fig:system_architecture}
\end{figure*}
%%%%%%%%%%%%%%%%%%%%%%%%%%%%%%%%%%%%%%%%%%%%%%%%%%%%%%%%%%%%%%%%%%%%%%%%%%%%%%%%
%%%%%%%%%%%%%%%%%%%%%%%%%%%%%%%%%%%%%%%%%%%%%%%%%%%%%%%%%%%%%%%%%%%%%%%%%%%%%%%%
\section{Method}

To address vision-based quadrotor navigation in dynamic environments under partial observability, we propose an anticipatory risk-guided reinforcement learning framework (Fig.~\ref{fig:system_architecture}). Specifically, a spatio-temporal encoder extracts motion cues directly from onboard depth sequences. During training, an asymmetric actor-critic architecture leverages privileged simulator states to construct dense future risk maps for explicit representation supervision. This cohesive design enables the deployed network to self-predict these collision risks, transferring anticipatory capabilities into a purely vision-based policy for robust dynamic avoidance without explicit object tracking or inference-time overhead.

\subsection{Problem Formulation and Observation Design}
\label{sec:Problem Formulation and Observations}

We formulate the vision-based dynamic navigation task as a Partially Observable Markov Decision Process (POMDP), defined by the tuple $\langle \mathcal{S}, \mathcal{A}, \mathcal{P}, \mathcal{R}, \Omega, \mathcal{O} \rangle$. Here, the true system state $s_t \in \mathcal{S}$ encompasses the complete kinematics of the quadrotor and the exact poses and velocities of all dynamic obstacles. Since $s_t$ is hidden during real-world deployment, the agent interacts with the environment through an onboard observation $\mathbf{o}_t \in \Omega$, which is generated according to the observation probability distribution $\mathcal{O}(\mathbf{o}_t \mid s_t)$. The objective is to learn an optimal policy $\pi^*$ that maximizes the expected cumulative discounted reward:
\begin{equation}
\pi^* = \arg\max_{\pi} \; \mathbb{E}_{\pi} \left[ \sum_{t=0}^{\infty} \gamma^t R(s_t, a_t) \right].
\end{equation}

\subsubsection{Observation Space}

At time step $t$, the onboard observation is formally defined as $\mathbf{o}_t = \left( \mathcal{L}_t,\; \mathbf{e}_t \right) \in \Omega$, where $\mathcal{L}_t$ captures the spatio-temporal perceptual history and $\mathbf{e}_t$ represents the proprioceptive state.

Raw depth images encode distances along the camera's optical axis, which are suboptimal for ego-centric spatial reasoning and collision checking due to peripheral distortion. To obtain a geometry-consistent representation, the raw depth image is first processed by a $3\times3$ minimum pooling operation to yield a conservatively filtered map $\mathcal{D}_t^{\min} \in \mathbb{R}^{H \times W}$. We then discretize the camera's valid field of view into $N_h = 34$ azimuth ($\theta_i$) and $N_v = 6$ elevation ($\phi_j$) bins, sampling the corresponding depth values $\tilde{D}_{t,ij}$ directly from $\mathcal{D}_t^{\min}$.

\begin{figure}[htbp]
    \centering
    \includegraphics[width=1\linewidth, trim=1 1 1 2, clip]{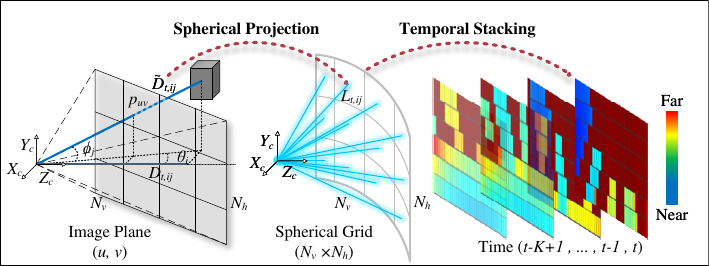}
    \caption{Illustration of the spatio-temporal observation space. Raw depth measurements are projected onto a geometry-consistent inverted spherical grid to mitigate optical-axis distortion and prioritize proximal collision risks. The sequential range maps are subsequently concatenated along the temporal dimension to explicitly encode the kinematics of dynamic obstacles.}
    \label{fig:observation_space}
\end{figure}

As illustrated in Fig.~\ref{fig:observation_space}, to convert these optical-axis depths into true Euclidean distances and emphasize imminent collision threats, we construct an inverted spherical range map $L_t \in \mathbb{R}^{N_v \times N_h}$. Each element is computed as:
% \begin{equation}
% L_{t,ij} = \max \left(0, \; d_{\max} - \tilde{D}_{t,ij} \cdot \sqrt{1 + \tan^2(\theta_i) + \tan^2(\phi_j)} \right),
% \end{equation}
\begin{equation}
\small
L_{t,ij}=\max\left(
0,d_{\max}-\tilde{D}_{t,ij}
\sqrt{1+\tan^2(\theta_i)+\tan^2(\phi_j)}
\right).
\end{equation}

where $d_{\max}$ denotes the maximum reliable sensing range. To ensure physical validity and robustness against real-world sensor artifacts, invalid pixels are systematically recovered via spatial interpolation prior to this transformation, preserving local geometric continuity.

To encode short-term motion context and resolve dynamic ambiguity, we maintain a temporal stack of the $K = 15$ most recent range maps:
\begin{equation}
\mathcal{L}_t = \left[ L_{t-K+1},\; \dots,\; L_t \right] \in \mathbb{R}^{K \times N_v \times N_h}.
\end{equation}

This compact representation minimizes real-time inference overhead while effectively capturing essential motion cues. Crucially, by abstracting away high-frequency sensor noise and domain-specific artifacts, it bridges the reality gap to facilitate robust zero-shot Sim-to-Real transfer.

Finally, the proprioceptive state is defined as $\mathbf{e}_t = [\mathbf{p}_{\text{rel}},\; \mathbf{v}_t]$, where $\mathbf{p}_{\text{rel}}$ is the relative goal position, $\mathbf{v}_t$ is the quadrotor's translational velocity, all expressed in the body frame.

\subsubsection{Action Space}
The action space $\mathcal{A} \subset \mathbb{R}^3$ specifies the desired translational velocity in the quadrotor's body frame:
\begin{equation}
a_t = [v_x,\; v_y,\; v_z]^\top \in [-v_{\max}, v_{\max}]^3.
\end{equation}
Operating directly at the velocity level abstracts away complex, hardware-specific rotor dynamics, thereby simplifying the learning process and significantly improving the robustness of zero-shot Sim-to-Real transfer. 

\subsection{Privileged Future Collision Risk Modeling}
\label{sec:future_risk}

To guide the extraction of motion-aware latent representations, we formulate a directionally aligned \emph{future collision risk} based on the Closest Point of Approach (CPA). Rather than computing this metric analytically during online execution, we utilize it exclusively in simulation as a dense, privileged supervisory signal to train the visual encoder.

All computations are performed in an ego-centric body frame. Let $\mathbf{p}_k = \mathbf{p}_{\mathrm{obs},k} - \mathbf{p}_{\mathrm{drone}}$ and $\mathbf{v}_k = \mathbf{v}_{\mathrm{obs},k} - \mathbf{v}_{\mathrm{drone}}$ denote the relative position and velocity of the $k$-th obstacle. Assuming an approximately constant relative velocity over a short prediction horizon $T$, the theoretical time of closest approach is analytically given by:
\begin{equation}
t_k^* = -\frac{\mathbf{p}_k^\top \mathbf{v}_k}{\|\mathbf{v}_k\|^2 + \epsilon},
\end{equation}
where $\epsilon = 10^{-5}$ is a small constant added for numerical stability. We bound the effective evaluation time to the prediction horizon as $\hat{t}_k = \min\bigl( \max(t_k^*, 0),\; T \bigr)$. The minimum separation distance within this horizon is thus defined as $d_k^* = \| \mathbf{p}_k + \hat{t}_k \mathbf{v}_k \|$. 

\begin{figure}[htbp]
    \centering
    \includegraphics[width=1\linewidth, trim=1 1 1 1, clip]{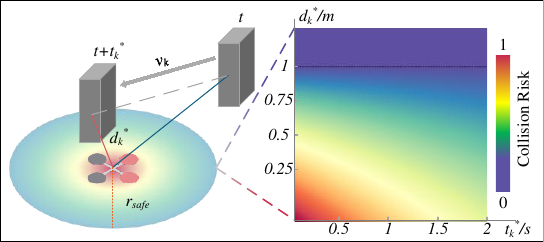}
    \caption{Visualization of the privileged future collision risk. The figure illustrates the kinematic geometry of the CPA in the ego-centric frame alongside the proposed bivariate risk function $R_{\mathrm{future}}^{(k)}$, which maps temporal urgency and spatial proximity to a continuous collision hazard score.}
    \label{fig:cpa_risk}
\end{figure}

As visualized in Fig.~\ref{fig:cpa_risk}, we formulate the future risk as a bivariate function of time-to-CPA ($\hat{t}_k$) and spatial proximity ($d_k^*$), which explicitly differentiates direct collisions from near-misses. For a hazardous obstacle ($\hat{t}_k \ge 0 \text{ and } d_k^* < r_{\mathrm{safe}}$), the continuous risk is defined as:
\begin{equation}
R_{\mathrm{future}}^{(k)} =
\begin{cases}
e^{-\alpha \hat{t}_k} \left( 1 - \frac{d_k^*}{r_{\mathrm{safe}}} \right)^\beta, & \hat{t}_k \ge 0, \, d_k^* < r_{\mathrm{safe}} \\
0, & \text{otherwise}
\end{cases}
\end{equation}
where $r_{\mathrm{safe}} = r_{\mathrm{drone}} + r_{\mathrm{obs}} + d_{\mathrm{buffer}}$ is the predefined safety margin explicitly accounting for geometric radii and a safety buffer. The parameter $\alpha > 0$ is a temporal scaling factor, while $\beta \ge 1$ controls the spatial decay shape. We empirically set $\beta = 2$ to ensure $C^1$ smoothness at the safety boundary $r_{\mathrm{safe}}$, thereby eliminating gradient discontinuities during training.

Furthermore, we construct a 2D spherical risk map $\mathbf{M} \in \mathbb{R}^{N_v \times N_h}$ that matches the sensor's angular resolution, ensuring the privileged risk is strictly aligned with the perception space. Each ray $(\theta_i, \phi_j)$ is assigned the maximum risk $R_{\mathrm{future}}^{(k)}$ among all obstacles whose safety spheres ($r_{\mathrm{safe}}$) intersect it. 

Finally, we compress this map into a 1D azimuthal vector $\mathbf{R} \in \mathbb{R}^{N_h}$ using a Log-Sum-Exp (LSE) smooth maximum pooling over the elevation dimension:
\begin{equation}
R_i = \frac{1}{\tau} \log \sum_{j=1}^{N_v} \exp(\tau {M}_{j,i}),
\end{equation}
where $\tau > 0$ is a temperature parameter controlling the approximation smoothness. This strictly differentiable LSE mechanism effectively preserves gradient flows to sub-optimal hazards and inherently encodes obstacle density by penalizing tightly clustered threats. Furthermore, compressing the risk into a 1D azimuthal space naturally filters out vertically safe obstacles and explicitly guides the policy toward horizontal escape routes---the dominant avoidance behavior for multi-rotors---while significantly reducing the dimensionality of the supervisory signal $\mathbf{R}$.

\subsection{Risk-Guided Spatio-Temporal Representation}
\label{sec:spatiotemporal_depth}

We propose a decoupled spatio-temporal encoding framework designed to extract motion-aware latent features from sequential depth observations $\mathcal{L}_t \in \mathbb{R}^{K \times N_v \times N_h}$, where $K$ denotes the temporal window size. The encoding process separates spatial and temporal dependencies: each frame ${L}_{t,k} \in \mathcal{L}_t$ is first processed by a shared 2D CNN to yield a spatial embedding $\mathbf{s}_{t,k} = \Phi_{\text{cnn}}({L}_{t,k})$, which is subsequently aggregated by a Temporal Convolutional Network (TCN) to produce the spatio-temporal feature vector $\mathbf{h}_t \in \mathbb{R}^{d_s}$:
\begin{equation}
\mathbf{h}_t = \Psi_{\text{tcn}}(\mathbf{s}_{t,1}, \dots, \mathbf{s}_{t,K}).
\end{equation}

To ensure that $\mathbf{h}_t$ captures task-relevant interaction dynamics, an auxiliary prediction head decodes the latent vector to estimate a 1D future collision risk map $\mathbf{\hat{R}}_t = f_{\text{aux}}(\mathbf{h}_t) \in [0, 1]^{N_h}$. This prediction is supervised by the privileged risk $\mathbf{R}_t$ derived in Sec.~\ref{sec:future_risk}, forcing the latent space to prioritize motion-consistent structures. Concurrently, $\mathbf{h}_t$ is fused with the proprioceptive state $\mathbf{e}_t$ via an MLP to form the backbone representation $\mathbf{z}_t = \text{MLP}([\mathbf{h}_t, \mathbf{e}_t])$. 
The final input $\mathbf{s}_t^A$ for the actor network is constructed as:
\begin{equation}
\mathbf{s}_t^A = [\mathbf{z}_t, \text{sg}(\mathbf{\hat{R}}_t)],
\end{equation}
where $\text{sg}[\cdot]$ denotes the stop-gradient operator. While the spatio-temporal backbone remains end-to-end trainable to adapt to navigation objectives, the detachment of $\mathbf{\hat{R}}_t$ prevents high-variance policy gradients from corrupting the auxiliary head. This ensures that the risk map provides an objective, semantically-consistent prior for decision-making.

\subsection{Asymmetric Risk-Guided RL Framework}
\label{sec:asym_rl}

We train the anticipatory navigation policy within an asymmetric actor--critic framework, leveraging privileged simulation data to stabilize learning and mitigate the partial observability inherent in depth-based dynamic navigation.

\paragraph{Asymmetric Architecture}
The actor $\pi_\theta(\mathbf{a}_t | \mathbf{s}_t^A)$ operates exclusively on the vision-based representation $\mathbf{s}_t^A$ defined in Sec.~\ref{sec:spatiotemporal_depth}. To provide accurate value estimation during training, the critic $V_\phi(\mathbf{s}_t^C)$ is conditioned on an augmented state $\mathbf{s}_t^C = [\mathbf{s}_t^A, \mathbf{s}_t^{\text{priv}}]$. The privileged component $\mathbf{s}_t^{\text{priv}}$ encompasses the ground-truth kinematics of the quadrotor and all dynamic obstacles provided directly by the simulator.  This configuration enables high-fidelity value estimation and variance reduction during training, while ensuring the actor remains strictly realizable using only onboard sensors during inference.

\paragraph{Reward Function}
To balance aggressive goal-seeking with absolute safety, the total reward $r_t$ is formulated as a weighted sum of navigation progress and associated operational costs:
\begin{equation}
r_t = w_n r_t^{\text{nav}} - w_s p_t^{\text{safe}} - w_r p_t^{\text{reg}} ,
\end{equation}
where $w_n$, $w_s$, and $w_r$ are positive weighting coefficients.

The navigation term $r_t^{\text{nav}}$ incentivizes continuous progress toward the goal:
\begin{equation}
r_t^{\mathrm{nav}} =
\alpha_d (d_{t-1} - d_t)
+ \alpha_v \mathbf{v}_t^\top \hat{\mathbf{g}}_t
+ \mathbb{I}_{\mathrm{success}} ,
\end{equation}
where $d_t$ is the Euclidean distance to the goal, $\hat{\mathbf{g}}_t$ is the unit vector pointing toward the target, and $\mathbb{I}_{\text{success}}$ is a sparse bonus granted upon mission completion.

Absolute safety is enforced via the penalty term $p_t^{\text{safe}}$, which explicitly quantifies both instantaneous spatial proximity and future collision threats:
\begin{equation}
p_t^{\mathrm{safe}} =
\exp(-\beta_s d_t^{\min})
+ \gamma_s \max_i R_{t,i} ,
\end{equation}
where $d_t^{\min}$ is the minimum instantaneous distance to any obstacle within the field of view (regardless of whether it is static or dynamic), $\beta_s$ is a sensitivity decay factor, and $R_{t,i}$ represents the elements of the privileged future risk map $\mathbf{R}_t$. The first term imposes a strict exponential cost for encroaching on any immediate physical boundaries, guaranteeing instantaneous geometric clearance. In contrast, the second term provides a dense, anticipatory penalty specifically targeting future dynamic collision risks, explicitly encouraging preemptive avoidance maneuvers.

The regularization penalty $p_t^{\text{reg}}$ ensures trajectory smoothness and physical feasibility:
\begin{equation}
p_t^{\mathrm{reg}} =
\eta_a \|\mathbf{v}_t - \mathbf{v}_{t-1}\|^2
+ \eta_j \|\mathbf{v}_t - 2\mathbf{v}_{t-1} + \mathbf{v}_{t-2}\|^2 ,
\end{equation}
These high-order differential terms minimize acceleration and jerk fluctuations, bridging the Sim-to-Real gap by penalizing erratic control commands.

\begin{figure}[htbp]
    \centering
    \includegraphics[width=1\linewidth]{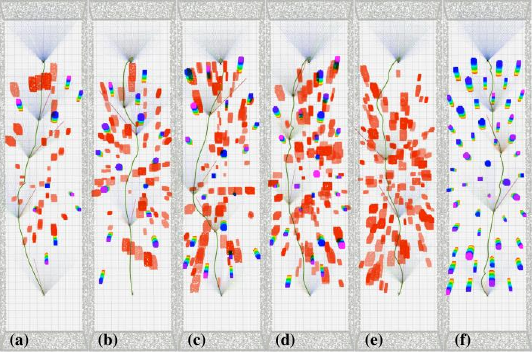} % 请确认你的文件名
    \caption{Top-down view of the UAV trajectories in the simulated benchmark environments under varying obstacle densities and dynamic configurations (a--f), which correspond to the six evaluation scenarios in Table~\ref{tab:Comparison} in sequence.}
    \label{fig:sim_trajectories}
\end{figure}

\paragraph{Training Procedure}
The policy is optimized using Proximal Policy Optimization (PPO) \cite{schulman2017proximal} within the NVIDIA Isaac Sim environment, building upon the massively parallel training framework established in OmniDrones \cite{xu2024omnidrones}. To regularize the shared spatio-temporal representation, the joint objective function incorporates the auxiliary risk supervision:
\begin{equation}
\mathcal{L} = \mathcal{L}_{\mathrm{PPO}} + \lambda_r \, \mathcal{L}_{\mathrm{risk}} .
\end{equation}

where $\mathcal{L}_{\text{risk}} = \frac{1}{N_h} \| \mathbf{\hat{R}}_t - \mathbf{R}_t \|^2$ is the Mean Squared Error (MSE) of the risk prediction. During real-world inference, the asymmetric critic and the ground-truth risk computation are discarded. Crucially, the trained auxiliary head is retained to explicitly self-predict the collision risk map $\mathbf{\hat{R}}_t$. This decoupled design ensures the policy operates exclusively on onboard spatial observations and its internalized anticipatory risk, enabling fully autonomous, zero-shot deployability.
%%%%%%%%%%%%%%%%%%%%%%%%%%%%%%%%%%%%%%%%%%%%%%%%%%%%%%%%%%%%%%%%%%%%%%%%%%%%%%%%
%%%%%%%%%%%%%%%%%%%%%%%%%%%%%%%%%%%%%%%%%%%%%%%%%%%%%%%%%%%%%%%%%%%%%%%%%%%%%%%%
\section{Experiment}

\subsection{Simulation Experiments}

\subsubsection{Simulation Setup}
Six simulated benchmark environments (Fig.~\ref{fig:sim_trajectories}), representing $30\,\mathrm{m} \times 10\,\mathrm{m}$ corridors cluttered with static and dynamic obstacles, are designed for systematic evaluation. Within these scenarios, dynamic obstacles execute back-and-forth linear motions between randomly sampled waypoints, with velocity magnitudes uniformly drawn from $[1, 5]\,\mathrm{m/s}$. For each environment, we conduct 20 independent trials featuring randomized goal locations and obstacle kinematics while maintaining a fixed starting position for the ego-agent. Success is defined as the quadrotor reaching the goal while maintaining a collision-free trajectory. Computational experiments are performed on a workstation equipped with an Intel i7-13700KF CPU and an NVIDIA RTX 3090 Ti GPU.

\begin{table*}[htbp]
\centering
\caption{\MakeUppercase{Benchmark results in different scenarios.}}
\label{tab:Comparison}

\setlength{\aboverulesep}{0pt}    % 消除横线上方的白边
\setlength{\belowrulesep}{0pt}    % 消除横线下方的白边
\setlength{\extrarowheight}{2pt} % 稍微增加一点行高，防止文字挤在横线上

% --- 颜色定义更新 ---
% 次优 浅红
\definecolor{cellsecondC}{rgb}{0.99, 0.92, 0.95} 
% 最优
\definecolor{cellbestC}{rgb}{0.94, 0.76, 0.80}   

% 定义命令：使用 cellcolor 填满单元格
\newcommand{\cbest}[1]{\cellcolor{cellbestC}\textbf{#1}}
\newcommand{\csecond}[1]{\cellcolor{cellsecondC}#1}

\resizebox{\textwidth}{!}{%
\setlength{\tabcolsep}{3pt}
\renewcommand{\arraystretch}{1.2}
\begin{tabular}{@{} l cccccc cccccc cccccc @{}}
\toprule
\multirow{2}{*}{\textbf{Method}} & \multicolumn{6}{c}{\textbf{10 static, 10 dynamic}} & \multicolumn{6}{c}{\textbf{15 static, 15 dynamic}} & \multicolumn{6}{c}{\textbf{20 static, 20 dynamic}} \\
\cmidrule(lr){2-7} \cmidrule(lr){8-13} \cmidrule(lr){14-19}
& \textbf{$\eta$} & \textbf{$t_p$} & \textbf{$v_a$} & \textbf{$R_l$} & \textbf{$d_a$} & \textbf{$d_m$} & \textbf{$\eta$} & \textbf{$t_p$} & \textbf{$v_a$} & \textbf{$R_l$} & \textbf{$d_a$} & \textbf{$d_m$} & \textbf{$\eta$} & \textbf{$t_p$} & \textbf{$v_a$} & \textbf{$R_l$} & \textbf{$d_a$} & \textbf{$d_m$} \\
\midrule
\cite{lu2025fapp} & 80 & 11.38 & 1.80 & \csecond{1.11} & \csecond{2.44} & 0.13 & 60 & 14.32 & 1.93 & \cbest{1.12} & \csecond{2.38} & 0.17 & 30 & 15.44 & 1.92 & 1.26 & \csecond{2.20} & 0.18 \\ 
\cite{xu2025navrl} & 70 & 17.92 & 1.41 & 1.20 & 1.48 & 0.15 & 60 & 20.51 & 1.25 & 1.31 & 1.42 & \csecond{0.20} & 40 & 18.90 & 1.50 & \csecond{1.21} & 0.98 & 0.12 \\
\cite{xu2025flow} & \csecond{90} & \csecond{5.73} & \cbest{4.08} & 1.19 & 2.40 & \csecond{0.33} & \csecond{70} & \csecond{6.10} & \cbest{4.15} & \csecond{1.14} & 1.96 & 0.14 & \csecond{60} & \csecond{5.93} & \cbest{3.89} & 1.31 & 1.82 & \csecond{0.28} \\
\midrule
Ours & \cbest{95} & \cbest{3.39} & \csecond{4.04} & \cbest{1.09} & \cbest{3.06} & \cbest{0.43} & \cbest{80} & \cbest{3.24} & \csecond{3.61} & 1.18 & \cbest{2.89} & \cbest{0.43} & \cbest{65} & \cbest{3.83} & \csecond{3.49} & \cbest{1.15} & \cbest{2.64} & \cbest{0.40} \\

\midrule
\addlinespace[0.5em]
\midrule

\multirow{2}{*}{\textbf{Method}} & \multicolumn{6}{c}{\textbf{25 static, 25 dynamic}} & \multicolumn{6}{c}{\textbf{35 dynamic}} & \multicolumn{6}{c}{\textbf{50 static}} \\
\cmidrule(lr){2-7} \cmidrule(lr){8-13} \cmidrule(lr){14-19}
& \textbf{$\eta$} & \textbf{$t_p$} & \textbf{$v_a$} & \textbf{$R_l$} & \textbf{$d_a$} & \textbf{$d_m$} & \textbf{$\eta$} & \textbf{$t_p$} & \textbf{$v_a$} & \textbf{$R_l$} & \textbf{$d_a$} & \textbf{$d_m$} & \textbf{$\eta$} & \textbf{$t_p$} & \textbf{$v_a$} & \textbf{$R_l$} & \textbf{$d_a$} & \textbf{$d_m$} \\
\midrule
\cite{lu2025fapp} & 20 & 18.16 & 1.68 & 1.34 & \csecond{2.10} & \csecond{0.18} & 20 & 16.76 & 1.95 & \cbest{1.07} & \csecond{1.86} & \csecond{0.19} & 70 & 17.37 & 1.85 & \cbest{1.12} & \csecond{1.23} & \csecond{0.15} \\ 
\cite{xu2025navrl} & 15 & 23.30 & 1.21 & 1.31 & 0.70 & 0.10 & 20 & 20.21 & 1.28 & 1.44 & 0.80 & 0.13 & \csecond{65} & 17.17 & 1.00 & 1.67 & 0.66 & 0.10 \\
\cite{xu2025flow} & \csecond{40} & \csecond{6.06} & \cbest{3.73} & \csecond{1.30} & 1.75 & 0.14 & \csecond{40} & \csecond{4.91} & \cbest{4.21} & \csecond{1.11} & 1.63 & 0.09 & \csecond{60} & \csecond{6.10} & \cbest{3.52} & 1.15 & 1.05 & 0.10 \\
\midrule
Ours & \cbest{55} & \cbest{3.23} & \csecond{3.40} & \cbest{1.15} & \cbest{2.82} & \cbest{0.35} & \cbest{50} & \cbest{4.07} & \csecond{3.15} & 1.13 & \cbest{2.72} & \cbest{0.31} & \cbest{70} & \cbest{3.22} & \csecond{2.71} & \csecond{1.14} & \cbest{2.19} & \cbest{0.43} \\
\bottomrule
\end{tabular}}
\end{table*}

Comprehensive assessment of navigation performance, safety, and computational efficiency is conducted via six quantitative metrics. These include \textbf{Success Rate} ($\eta$, \%), representing the percentage of collision-free trials; \textbf{Average Speed} ($v_a$, $\mathrm{m/s}$); \textbf{System Latency} ($t_p$, $\mathrm{ms}$), denoting the total end-to-end execution time from raw sensor input to action output; \textbf{Path Efficiency} ($R_l$), comparing the traveled distance against the Euclidean trajectory; \textbf{Average Clearance} ($d_a$, $\mathrm{m}$); and \textbf{Minimum Clearance} ($d_m$, $\mathrm{m}$), which serves as a strict metric for worst-case collision risk. Continuous metrics ($v_a$, $t_p$, $R_l$, $d_a$, $d_m$) are derived solely from successful trials.

\subsubsection{Comparison to Baselines}

We evaluate our proposed method against three representative frameworks, namely the optimization-based FAPP \cite{lu2025fapp}, the hierarchical RL method NavRL \cite{xu2025navrl}, and the point-flow-based P2M \cite{xu2025flow}. Proposed benchmarks ensure a rigorous comparison by deploying all baselines under their officially optimized kinematic configurations. Specifically, we restrict FAPP \cite{lu2025fapp} and NavRL \cite{xu2025navrl} to velocities of $3.5\,\mathrm{m/s}$ and $2\,\mathrm{m/s}$ respectively to preserve estimation stability and accommodate sensing limitations. In contrast, our method and P2M \cite{xu2025flow} both target a high-speed $5\,\mathrm{m/s}$ reference velocity. Notably, our framework achieves this performance under significantly stricter perception constraints by relying solely on a forward-facing depth sensor as opposed to the omnidirectional LiDAR necessitated by P2M \cite{xu2025flow}.

\begin{figure}[htbp]
    \centering
    \includegraphics[width=1\linewidth, trim=1 1 1 1, clip]{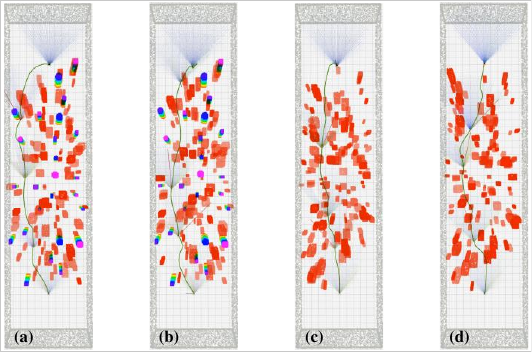}
    \caption{Ablation study on UAV flight trajectories in the highly cluttered dynamic scenario. (a) and (c) showcase the impact of removing risk map supervision, leading to aggressive near-collision behaviors. (b) and (d) demonstrate the policy without TCN, which results in reactive and unstable maneuvers.}
    \label{fig:ablation_trajectories}
\end{figure}

Table \ref{tab:Comparison} demonstrates the performance boundaries of the evaluated methods. Explicit perception pipelines like \cite{lu2025fapp} and \cite{xu2025navrl} suffer from high planning latency due to multi-target tracking and mapping overhead. Consequently, they are forced to adopt highly conservative flight behaviors, yielding sluggish average velocities. As obstacle density escalates, this systemic latency severely degrades their reactivity. Their success rates plummet, and their absolute minimum clearances drop to near-collision thresholds, indicating a profound failure to guarantee safety in highly chaotic clutters.

\begin{figure*}[htbp] 
    \centering
    \includegraphics[width=\textwidth]{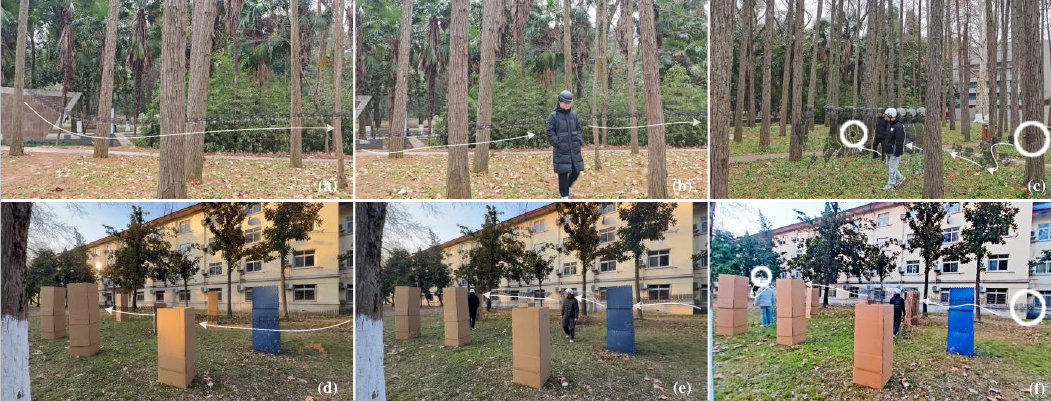} 
    \caption{Snapshots of autonomous UAV navigation across diverse real-world clutters. (a) and (d) demonstrate spatial generalization in an unstructured forest and a structured box environment. (b) through (e) illustrate anticipatory avoidance during dynamic pedestrian encounters. (f) highlights the agile reactive response of the policy to dynamic agents abruptly emerging from occluded regions.}
    \label{fig:real_experiments}
\end{figure*}

Conversely, \cite{xu2025flow} sustains high average velocities via omnidirectional LiDAR. However, optimizing complex spatio-temporal representations solely via aggregated scalar rewards presents a severe supervision bottleneck. Lacking explicit structural guidance, \cite{xu2025flow} fails to fully internalize dense motion dynamics, inevitably defaulting to reactive, late-stage evasions rather than proactive planning. This reactive behavior is clearly evidenced by its severely degraded minimum safety clearances across all scenarios. The policy avoids collisions solely through kinematically aggressive, near-collision detours, reflecting poor safety robustness in dense simulations.

Our method overcomes these limitations through future risk map supervision. This mechanism enables anticipatory planning without the computational overhead of explicit tracking. As demonstrated in the results, our approach consistently achieves the highest success rates across varying densities while strictly maintaining the largest minimum safety clearances. Furthermore, it guarantees low inference latency alongside competitive flight speeds. Ultimately, the proposed framework successfully balances high-speed flight efficiency and robust safety under limited sensing constraints.

\subsubsection{Ablation on Method Components}

We evaluate the individual contributions of the Temporal Convolutional Network and the auxiliary risk prediction module via an ablation study across two highly cluttered scenarios namely a mixed environment of static and dynamic obstacles and a purely dynamic environment as detailed in Table~\ref{tab:ablation}.

\begin{table}[htbp]
\centering
\caption{Ablation results across two highly cluttered scenarios.}
\label{tab:ablation}
\small
\setlength{\tabcolsep}{3pt}
\renewcommand{\arraystretch}{1.15}

% --- 颜色定义更新 ---
% 次优 浅红
\definecolor{cellsecondC}{rgb}{0.99, 0.92, 0.95} 
% 最优
\definecolor{cellbestC}{rgb}{0.94, 0.76, 0.80}   

\newcommand{\cbest}[1]{\colorbox{cellbestC}{#1}}
\newcommand{\csecond}[1]{\colorbox{cellsecondC}{#1}}

\begin{tabular}{@{} >{\centering\arraybackslash}p{1.6cm} c c c c c c @{}}
\toprule
\textbf{Scenario} & \textbf{Method} & {$\eta$}  & {$v_a$} & {$R_l$} & {$d_a$} & {$d_m$}\\
\midrule
\multirow{3}{*}{\shortstack[c]{25 static \\ 25 dynamic}}
& Ours      & \cbest{\textbf{55}} & \csecond{3.40}        & \cbest{\textbf{1.15}} & \cbest{\textbf{2.82}} &\cbest{\textbf{0.35}}\\
& w/o Risk  & {40}        & \cbest{\textbf{3.42}} & {1.18}        & {2.51} &\csecond{0.29}      \\
& w/o TCN   & \csecond{50}                  & 3.24                  & \csecond{1.16}                 & \csecond{2.79}   &0.26               \\
\midrule
\multirow{3}{*}{\shortstack[c]{35 dynamic}}
& Ours      & \cbest{\textbf{50}} & \csecond{3.15}        & \cbest{\textbf{1.13}} & \cbest{\textbf{2.72}} &\cbest{\textbf{0.31}}\\
& w/o Risk  & \csecond{30}        & \cbest{\textbf{3.30}} & \csecond{1.15}        & \csecond{2.68} &\csecond{0.23}      \\
& w/o TCN   & 20                  & 3.05                  & 1.19                  & 2.42   &\csecond{0.23}                \\
\bottomrule
\end{tabular}
\end{table}

Removing the temporal feature extractor (\textbf{w/o TCN}) reduces the framework to a memoryless spatial policy. Without the ability to track short-term motion, the agent struggles to capture the dynamic trends of the environment and fails to perceive the potential collision risks induced by relative motion. Lacking temporal context, the drone is forced into a purely reactive paradigm, executing sudden, late-stage evasions that severely compress its safety margins and lead to dangerous near-collisions, particularly in purely dynamic scenarios.

Eliminating the auxiliary risk module (\textbf{w/o Risk}) deprives the framework of structured geometric motion priors. Without explicitly predicting and conditioning on the future collision risk map, the policy must deduce complex dynamic interactions solely from implicit, task-level scalar rewards. As discussed, such scalar feedback lacks spatial and temporal structure. This deficiency is highly exposed in mixed clutters, where the agent fails to identify safety-critical motion cues and makes sub-optimal navigation decisions. Consequently, it resorts to aggressive, unsafe bypasses rather than anticipatory detours, trading essential safety boundaries for marginally higher flight speeds.

Ultimately, these results validate the proposed synergistic integration. The temporal network captures essential dynamic trends, while the self-predicted risk map explicitly translates these trends into anticipatory collision awareness. Together, they successfully prioritize absolute safety over reckless agility during high-speed flight.

\subsection{Real-World Experiments}
\label{sec:experiments}

\subsubsection{Hardware Deployment and Sim-to-Real Transfer}
\label{subsec:hardware}
Our vision-based quadrotor utilizes an NVIDIA Jetson Orin NX for VINS state estimation \cite{qin2018vins} and policy inference, alongside an Intel RealSense D435i for depth and stereo imaging. To facilitate zero-shot Sim-to-Real transfer, we leverage the \textit{representation abstraction} detailed in Sec.~\ref{sec:Problem Formulation and Observations}. By filtering raw sensor artifacts into structured geometric abstractions, this multi-stage processing ensures that noisy real-world depth inputs strictly align with the idealized observations used during simulation training, effectively bridging the reality gap without domain randomization.

TensorRT acceleration minimizes policy inference latency to approximately $5\,\mathrm{ms}$, thereby ensuring closed-loop stability and mitigating physical execution delays. The proposed architecture subsequently leverages a PX4-based flight controller to track the continuous velocity commands at high frequency.

\subsubsection{Navigation in Static Environments}
\label{subsec:static_nav}
We evaluate spatial generalization and agile maneuverability under perception constraints via autonomous navigation trials in an unstructured outdoor forest and a structured indoor environment comprising unevenly stacked boxes. The UAV seamlessly generates smooth avoidance maneuvers upon perceiving trees and box geometries as depicted in Fig.~\ref{fig:real_experiments} a and d. Robust and collision-free commands are maintained despite encountering textures and structural distributions entirely unseen during simulation training. This performance validates the excellent zero-shot Sim-to-Real transfer capability of the proposed representation.

\subsubsection{Navigation in Dynamic-Static Environments}
\label{subsec:dynamic_nav}
To further validate anticipatory avoidance against moving threats, we introduced pedestrians with complex, fast-moving trajectories into the static scenes. As illustrated in Fig.~\ref{fig:real_experiments}(b), (c), (e), and (f), the UAV effectively anticipates potential collisions induced by pedestrian movements and executes evasive maneuvers proactively. This confirms that the spatio-temporal encoder, explicitly supervised by the short-horizon future collision risk during training, successfully extracts kinematic cues to adjust the flight path before threats become imminent. Notably, in Fig.~\ref{fig:real_experiments}(e), when a pedestrian suddenly emerges from an occlusion blind spot, the UAV exhibits remarkable agility: it forcefully brakes to a rapid hover before executing a safe lateral detour. This behavior highlights the policy's capacity for limit-of-handling reactions and dynamic adaptability, underscoring the system's robustness in highly cluttered, human-robot hybrid environments.
%%%%%%%%%%%%%%%%%%%%%%%%%%%%%%%%%%%%%%%%%%%%%%%%%%%%%%%%%%%%%%%%%%%%%%%%%%%%%%%%
%%%%%%%%%%%%%%%%%%%%%%%%%%%%%%%%%%%%%%%%%%%%%%%%%%%%%%%%%%%%%%%%%%%%%%%%%%%%%%%%
\section{Conclusion}
\label{sec:conclusion}

We present a risk-guided reinforcement learning framework for anticipatory quadrotor navigation in highly dynamic environments. The proposed asymmetric actor-critic architecture utilizes privileged future collision risks from the Closest Point of Approach as explicit spatio-temporal feature guidance. This mechanism enables a lightweight encoder to learn motion-aware representations directly from sequences of inverted spherical depth maps. Experimental results validate that this methodology achieves robust zero-shot Sim-to-Real transfer and maintains superior safety margins in dense clutter by executing proactive maneuvers instead of aggressive reactive behaviors. Future research will address the current limitations regarding the constant-velocity assumption and forward-facing sensor blind spots by integrating learning-based trajectory forecasting and omnidirectional perception.
%%%%%%%%%%%%%%%%%%%%%%%%%%%%%%%%%%%%%%%%%%%%%%%%%%%%%%%%%%%%%%%%%%%%%%%%%%%%%%%%

\addtolength{\textheight}{-12cm}   % This command serves to balance the column lengths

%%%%%%%%%%%%%%%%%%%%%%%%%%%%%%%%%%%%%%%%%%%%%%%%%%%%%%%%%%%%%%%%%%%%%%%%%%%%%%%%
% \begin{thebibliography}{99}
% \bibitem{c1} G. O. Young, ÒSynthetic structure of industrial plastics (Book style with paper title and editor),Ó 	in Plastics, 2nd ed. vol. 3, J. Peters, Ed.  New York: McGraw-Hill, 1964, pp. 15Ð64.
% \bibitem{c2} W.-K. Chen, Linear Networks and Systems (Book style).	Belmont, CA: Wadsworth, 1993, pp. 123Ð135.
% \bibitem{c3} H. Poor, An Introduction to Signal Detection and Estimation.   New York: Springer-Verlag, 1985, ch. 4.
% \end{thebibliography}

\end{document}